\documentclass[runningheads]{llncs}
\usepackage{graphicx}
\usepackage{amsmath,amssymb} 
\usepackage{color}

\usepackage{float}
\usepackage[width=122mm,left=12mm,paperwidth=146mm,height=193mm,top=12mm,paperheight=217mm]{geometry}
\usepackage{wrapfig}

\usepackage{hyperref}       
\usepackage{url}            
\usepackage{booktabs}       
\usepackage{nicefrac}       
\usepackage{microtype}      
\usepackage[normalem]{ulem}
\usepackage{algorithm}
\usepackage{algorithmic}
\usepackage{epsfig}
\usepackage{graphicx}
\usepackage{color}
\usepackage{multirow} 
\usepackage{xspace}
\usepackage{capt-of}
\usepackage{enumitem}
\DeclareMathAlphabet{\pazocal}{OMS}{zplm}{m}{n}

\usepackage{subcaption}
\usepackage{float}

\def\maskconnect{{{MaskConnect}}\xspace}

\newcommand{\paramMillion}[1]{#1\texttt{M}}

\begin{document}

\title{MaskConnect: Connectivity Learning by Gradient Descent}
\titlerunning{MaskConnect: Connectivity Learning by Gradient Descent}

\authorrunning{Karim Ahmed, Lorenzo Torresani}

\author{Karim Ahmed \and Lorenzo Torresani}
\institute{Department of Computer Science, Dartmouth College\\
	\email{karim@cs.dartmouth.edu, LT@dartmouth.edu}
}

\maketitle

\begin{abstract}
Although deep networks have recently emerged as the model of choice for many computer vision problems, in order to yield good results they often require time-consuming architecture search. To combat the complexity of design choices, prior work has adopted the principle of modularized design which consists in defining the network in terms of a composition of topologically identical or similar building blocks (a.k.a. modules). This reduces architecture search to the problem of determining the number of modules to compose and how to connect such modules. Again, for reasons of design complexity and training cost, previous approaches have relied on simple rules of connectivity, e.g., connecting each module to only the immediately preceding module or perhaps to all of the previous ones. Such simple connectivity rules are unlikely to yield the optimal architecture for the given problem.

In this work we remove these predefined choices and propose an algorithm to learn the connections between modules in the network. Instead of being chosen a priori by the human designer, the connectivity is learned simultaneously with the weights of the network by optimizing the loss function of the end task using a modified version of gradient descent. We demonstrate our connectivity learning method on the problem of multi-class image classification using two popular architectures: ResNet and ResNeXt. Experiments on four different datasets show that connectivity learning using our approach yields consistently higher accuracy compared to relying on traditional predefined rules of connectivity. Furthermore, in certain settings it leads to significant savings in number of parameters.
\end{abstract}

\section{Introduction}
\label{sec:intro}

Deep neural networks have emerged as one of the most prominent models for problems that require the learning of complex functions and that involve large amounts of training data. While deep learning has recently enabled dramatic performance improvements in many application domains, the design of deep architectures is still a challenging and time-consuming endeavor. The difficulty lies in the many architecture choices that impact{---}often significantly{---}the performance of the system. In the specific domain of image categorization, which is the focus of this paper, significant research effort has been invested in the empirical study of how depth, filter sizes, number of feature maps, and choice of nonlinearities affect performance~\cite{GlorotEtAl:AISTATS2011,AlexNet,sermanet2013overfeat,MaasEtAl:ICML2013,ZeilerF14,GoogleLeNet}. Recently, several authors have proposed to simplify the architecture design by defining convolutional neural networks (CNNs) in terms of composition of topologically identical or similar building blocks or {\em modules}. This strategy was arguably first popularized by the VGG nets~\cite{SimonyanZisserman:ICLR2015} which were built by stacking a series of convolutional layers having identical filter size ($3 \times 3$). Other examples are ResNets~\cite{he2016deep} which are constructed by stacking residual blocks of fixed topology, ResNeXt models~\cite{XieGDTH16} which use multi-branch residual block modules, DenseNets~\cite{DenseNets} which use dense blocks as building blocks, or Multi-Fiber networks~\cite{chen2018multifiber} which use parallel branches (``fibers'') connected by routers (``transistors'').

While the principle of modularized design has greatly simplified the challenge of building effective architectures for image analysis, the choice of how to combine and aggregate the computations of these building blocks still rests on the shoulders of the human designer. To avoid a combinatorial explosion of options, prior work has relied on simple, uniform rules of aggregation and composition. For example, in ResNets and DenseNets each building block is connected only to the preceding one, via identity mapping, convolution or pooling. ResNeXt models~\cite{XieGDTH16} use a set of simplifying assumptions: the branching factor $C$ (also referred to as {\em cardinality}) is fixed to the same constant in all layers of the network, all branches of a module are fed the same input, and the outputs of parallel branches are aggregated by a simple additive operation that provides the input to the next module. While these simple rules of connectivity render network design more manageable, they are unlikely to yield the optimal connectivity for the given problem. 


In this paper we remove these predefined choices and propose an algorithm that learns to combine and aggregate building blocks of a neural network by directly optimizing connectivity of modules with respect to the given task. In this new regime, the network connectivity naturally arises as a result of training rather than being hand-defined by the human designer. 
While in principle this involves a search over an exponential number of connectivity configurations, our method can efficiently optimize the training loss with respect to connectivity using a variant of backpropagation. This is achieved by means of {\em connectivity masks}, i.e., learned binary parameters that act as ``switches'' determining the final connectivity in our network. The masks are learned together with the convolutional weights of the network, as part of a joint optimization with respect to the given loss function for the problem.

We evaluate our method on the problem of multi-class image classification using two popular modular architectures: ResNet and ResNeXt. We demonstrate that models with our learned connectivity consistently outperform the networks based on predefined rules of connectivity for the same budget of residual blocks (and parameters). An interesting byproduct of our approach is that, in certain settings, it can automatically identify modules that are superfluous, i.e., unnecessary or detrimental for the end objective. At the end of the optimization, these unused modules can be pruned away without  impacting the learned hypothesis while reducing substantially the runtime and the number of parameters to store.

By recasting the training procedure as an optimization over learning weights {\em and} connectivity, our method effectively searches over a larger space of solutions. This yields networks achieving higher accuracy than those constrained to use predefined connectivities. The average training time overhead is moderate, ranging between $13\%$ (for ResNet models) and  $39\%$ (for ResNeXt models) compared to learning using fixed connectivity which, however, yields lower accuracy. Finally we point out that, although our experiments are carried out using ResNet and RexNeXt models, our approach is general and applicable without major modifications to other forms of network architectures and other tasks beyond image categorization. In principle our method can also be used to learn connectivity among layers of a traditional (i.e., non-modular) neural network or a CNN. However, modern networks typically include a very large number of layers (hundreds or even thousands~\cite{eccv_HeZRS16}), which would make our approach very costly. Learning connectivity among modules is more manageable as each module encapsulates many layers and thus the total number of modules is typically small even for deep networks. 

\section{Related Work}

Despite their wide adoption, deep networks often require laborious model search in order to yield good results. As a result, significant research effort has been devoted to the design of algorithms for automatic model selection. However, most of this prior work falls within the genre of hyper-parameter optimization~\cite{BergstraBengio:JMLR2012,SnoekEtAl:NIPS2012,SnoekEtAl:ICML2015} rather than architecture or connectivity learning. Evolutionary search has been proposed as an interesting framework to learn both the structure as well as the connections in a neural network~\cite{pham2018efficient,such2017deep,salimans2017evolution,liu2017hierarchical,xie2017genetic,WierstraEtal:GECCO2005,Floreano:EI2008,EstebanReal:arXiv2017,PathNet}. Architecture search has also been recently formulated as a reinforcement learning problem with impressive results~\cite{ZophLe:ICLR2017}. 
Several authors have proposed learning connectivity by pruning unimportant weights from the network~\cite{LeCunDenkerSolla:NIPS90,han2015deep,han2015learning,GuoEtAl:NIPS16,han2016dsd}. However, these prior methods operate in stages where initially a network with full connectivity is learned and then connections are greedily removed according to an importance criterion. Compare to all these prior approaches, our work provides the advantage of learning the connectivity by direct global optimization of the loss function of the problem at hand rather than by greedy optimization of an auxiliary proxy criterion or by costly evolutionary search. Our technical approach shares similarities with the ``Shake-Shake'' regularization~\cite{shakeshake}. This procedure was demonstrated on two-branch ResNeXt models and consists in randomly scaling tensors produced by parallel branches during training while at test time the network uses uniform weighting of tensors. Conversely, our algorithm {\em learns} an optimal binary scaling of the parallel tensors with respect to the training objective and uses the resulting network with sparse connectivity at test time. 
While our algorithm is limited to optimizing the connectivity structure within a predefined architecture, Adams et al.~\cite{AdamsEtAl2010} proposed a nonparametric Bayesian approach  that searches over an infinite network using MCMC. Our approach can be viewed as a middle ground between two extremes: using hand-defined networks versus learning/searching the full architecture from scratch. The advantage is that our connectivity learning can be done without adding a significant training time overhead (only 13-39\% depending on the architecture) compared to using fixed connectivity. The disadvantage is that the space of models considered by our approach is a lot more constrained than in the case of general architecture search. Saxena and Verbeek~\cite{SaxenaVerbeek:NIPS2016} introduced convolutional neural fabric which are learnable 3D trellises that locally connect response maps at different layers of a CNN. Similarly to our work, they enable optimization over an exponentially large family of connectivities, albeit different from those considered here. Finally, our approach is also related to conditional computation methods~\cite{blockdrop,bengio2013estimating,bengio2015conditional,bengio2013deep,shazeer2017outrageously,davis2013low,eigen2013learning,denoyer2014deep,cho2014exponentially,almahairi2016dynamic}, which learn to drop out blocks of units. However, unlike these techniques, our algorithm learns a fixed, sparse connectivity that does {\em not} change with the input and thus it keeps the runtime cost and the number of used parameters constant. 
\section{Technical Approach}
\label{sec:approach}
\subsection{Modular architecture}
\label{sec:modular}

We begin by defining the modular architecture that will be used by our framework. In order to present our method in its full generality, we will describe it in the context of a {\em general} modular architecture, which we will then instantiate in the form of  the two models used in our experiments (ResNet and ResNeXt). 

We assume that the general modular architecture consists of a stack of $L$ modules. (When using ResNet the modules will be residual blocks, while for ResNeXt each module will consist of multiple parallel branches.) We denote with ${\bf x}_j$ the input to the $j$-th module for $j=1,\hdots,L$. The input of each module is an activation tensor computed from one the previous modules. We assume that the module implements a function $\mathcal{G}(.)$ parameterized by learnable weights $\theta_j$. The weights may for example represent the coefficients of convolutional filters. Thus, the output ${\bf y}_j$ computed by the $j$-th module is given by ${\bf y}_j = \mathcal{G}({\bf x}_j; \theta_j)$. In prior modular architectures, such as ResNet, ResNeXt and DenseNet, the connectivity between modules is hand-defined a priori according to a very simple rule: the input of a module is the output of the preceding module. In other words, 
${\bf x}_j \leftarrow {\bf y}_{j-1}~.$
While this makes network design straightforward, it greatly limits the topology of architectures considered for the given task. In the next subsection we describe how to parameterize the architecture to remove these constraints and to enable connectivity learning in modular networks.

\subsection{Masked architecture}
\label{sec:masked}

We now introduce learnable {\em masks} defining the connectivity in the network. Specifically, we want to allow each module $j$ to take input from one or more of the preceding modules $k=1,\hdots,j-1$. To achieve this we 
define for each module a binary mask vector that controls the input pathway of that module. The binary mask vectors are learned jointly with the weights of the network. Let {\small {${\bf m}_{j} = [{m}_{j,1}, {m}_{j,2}, \hdots, {m}_{j,j-1}]\smash{^\top} \in \{0, 1\}^{j-1}$}} be the binary mask vector defining the {\em active} input connections feeding the $j$-th module. If {\small \smash{${m}_{j,k} = 1$}}, then the activation volume produced by the $k$-th  module is fed as input to the $j$-th module. If {\small \smash{${m}_{j,k} = 0$}}, then the output from the $k$-th module is ignored by the $j$-th module. The tensors from the {\em active} input connections are all added together (in an element-wise fashion) to form the input to the module. Thus, if we denote again with {\small \smash{${\bf y}_k$}} the output activation tensor computed by the $k$-th module, the input {\small \smash{${\bf x}_j$}} to the $j$-th module will be given by the following equation:
\begin{equation} 
{\bf x}_j = \sum_{k=1}^{j-1} {m}_{j,k} \cdot {\bf y}_k
\label{eq:resinput}
\end{equation}
Then, the output of this module will be obtained through the usual computation, i.e., {\small \smash{${\bf y}_j = \mathcal{G}({\bf x}_j; \theta_j)$}}.
We note that under this model we no longer have predefined connectivity among modules. Instead, the mask {\small \smash{${\bf m}_{j}$}} now determines {\em selectively} for each module which outputs from the previous modules will be aggregated and form the input to the block. In this paper we constrain the aggregations of outputs from the active connections to be in the form of simple additions as this does not require new parameters. When different modules yield feature maps of different sizes, we use zero-padding shortcuts to increase the dimensions of feature tensors to the largest size (as in~\cite{he2016deep}). These shortcuts are parameter free. We leave to future work the investigation of more sophisticated, parameterized aggregation schemes.

We point out that depending on the constraints defined over {\small \smash{${\bf m}_{j}$}}, different interesting models can be realized. For example, by introducing the constraint that {\small \smash{$\sum_k {m}_{j,k} = 1$}} for each block $j$, then each module will receive input from only one of the preceding modules (since each {\small \smash{${m}_{j,k}$}} must be either 0 or 1). At the other end of the spectrum, if we set {\small \smash{${m}_{j,k}=1$}} for all modules $j,k$, then all connections would be active. In our experiments we will demonstrate that the best results are typically achieved for values in between these two extremes, i.e., by connecting each module to $K$ previous modules where $K$ is an integer-valued hyperparameter such that $1< K < (j-1)$. We refer to this hyperparameter as the {\em fan-in} of a module. As discussed in the next section, the mask vector {\small \smash{${\bf m}_{j}$}} for each block is learned simultaneously with all the other weights in the network via backpropagation. Finally, we note that it may be possible for a module in the network to become unused. This happens when, as a result of the optimization, module $k$ is such that {\small \smash{$m_{j,k}=0$ for all $j$}}. In this case, at the end of the optimization, we prune the module in order to reduce the number of parameters to store and to speed up inference (note that this does not affect the function computed by the network).
In the next subsection we discuss our method for jointly learning the weights and the masks in the network.

\subsection{\maskconnect: learning to connect}
\label{sec:masktraining}

We refer to our learning algorithm as \maskconnect. It performs joint optimization of a given learning objective $\ell$  with respect to both the weights of the network ($\theta$) as well as the masks (${\bf m}$). Since in this paper we apply our method to the problem of image categorization, we use the traditional multi-class cross-entropy objective for the loss $\ell$. However, our approach can be applied without change to other loss functions and other tasks benefitting from connectivity learning. 

In \maskconnect the weights have real values, as in traditional networks, while the masks have binary values. This renders the optimization  challenging. To learn these binary parameters, we adopt a modified version of backpropagation, inspired by the  algorithm proposed by Courbariaux et al.~\cite{NIPS2015_5647} to train neural networks with binary weights. During training we store and update a real-valued version {\small \smash{$\tilde{\bf m}_j \in \left[0,1\right]^{j-1}$}} of the masks, with entries clipped to lie between 0 and 1. 

In general, the training via backpropagation consists of three steps: 1) forward propagation, 2) backward propagation, and 3) parameters update. At each iteration, we stochastically binarize the real-valued masks into binary-valued vectors {\small \smash{${\bf m}_j \in \{0,1\}^{j-1}$}} which are then used for the forward propagation and backward propagation (steps 1 and 2). Instead, during the parameters update (step 3), the method updates the real-valued masks {\small \smash{$\tilde{\bf m}_j$}}. The weights $\theta$ of the convolutional and fully connected layers are optimized using standard backpropagation. We discuss below the details of our mask training procedure, under the constraint that at any time there can be only $K$ active entries in the binary mask {\small \smash{${\bf m}_j$}}, where $K$ is a predefined integer hyperparameter with $1\leq K \leq {j-1}$. In other words, we impose the following constraints: 
\begin{eqnarray}
 m_{j,k} \in \{0,1\} ~ \forall j,k, ~~~\text{and}~~~\sum_{k=1}^{j-1} m_{j,k} = K ~ \forall j.\nonumber
\end{eqnarray}
These constraints imply that each module receives input from exactly $K$ previous modules.  

\paragraph{Forward Propagation.} During the forward propagation, our algorithm first normalizes the real-valued  entries in the mask of each block $j$ to sum up to 1, such that {\small \smash{$\sum_{k=1}^{j-1}\tilde{m}_{j,k}=1$}}. This is done so that {\small \smash{$\text{Mult}(\tilde{m}_{j,1}, \tilde{m}_{j,2}, \hdots, \tilde{m}_{j,j-1})$}} defines a proper multinomial distribution over the $j-1$ possible input connections into module $j$. Then, the binary  mask {\small \smash{${\bf m}_{j}$}} is stochastically generated by drawing $K$ {\em distinct} samples  \smash{$a_1,a_2,\hdots,a_K \in \{1, \hdots, (j-1)\}$} from the multinomial distribution over the connections. 
Finally, the entries corresponding to the $K$ samples are activated in the binary mask vector, i.e., {\small \smash{$m_{j,a_k} \leftarrow 1$, for $k=1,...,K$}}. The input activation volume to the module $j$ is then computed according to Eq.~\ref{eq:resinput} from the sampled binary masks. We note that the sampling from the Multinomial distribution ensures that the connections with largest {\small \smash{$\tilde{m}_{j,k}$}} values will be more likely to be chosen, while at the same time the stochasticity of this process allows different connectivities to be explored, particularly during early stages of the learning when the real-valued masks still have fairly uniform distributions.

\paragraph{Backward Propagation.} In the backward propagation step, the gradient {\small \smash{${\partial \pazocal{\ell}}/{\partial y_{k}}$}} with respect to each output is obtained via back-propagation from {\small \smash{${\partial \pazocal{\ell}}/{\partial x_{j}}$}} and the binary masks {\small \smash{$m_{j,k}$}}.

\paragraph{Mask Update.} In the parameter update step our algorithm computes the gradient with respect to the binary masks for each module. Then, using these computed gradients and the given learning rate, it updates the real-valued masks via gradient descent. At this time we clip the updated real-valued masks to constrain them to remain within the valid interval $[0, 1]$ (as in~\cite{NIPS2015_5647}).

Pseudocode for our training procedure is given in Appendix~\ref{appendixA}. After joint training over $\theta$ and ${\bf m}$, we have found beneficial to (1) freeze the binary masks to the top-$K$ values for each mask (i.e., by setting as active connections in  {\small \smash{${{\bf m}}_{j}$}} those corresponding to the $K$ largest values in {\small \smash{$\tilde{{\bf m}}_{j}$}}) and then (2) fine-tune the weights $\theta$ of the network with respect to these fixed binary masks. 

In the next subsections we discuss how we instantiated our general approach for the two architectures considered in our experiments: ResNet and ResNeXt.

\subsection{\maskconnect applied to ResNet}

\begin{figure}[t!]
  \begin{center}
  \includegraphics[width=1.0\linewidth]{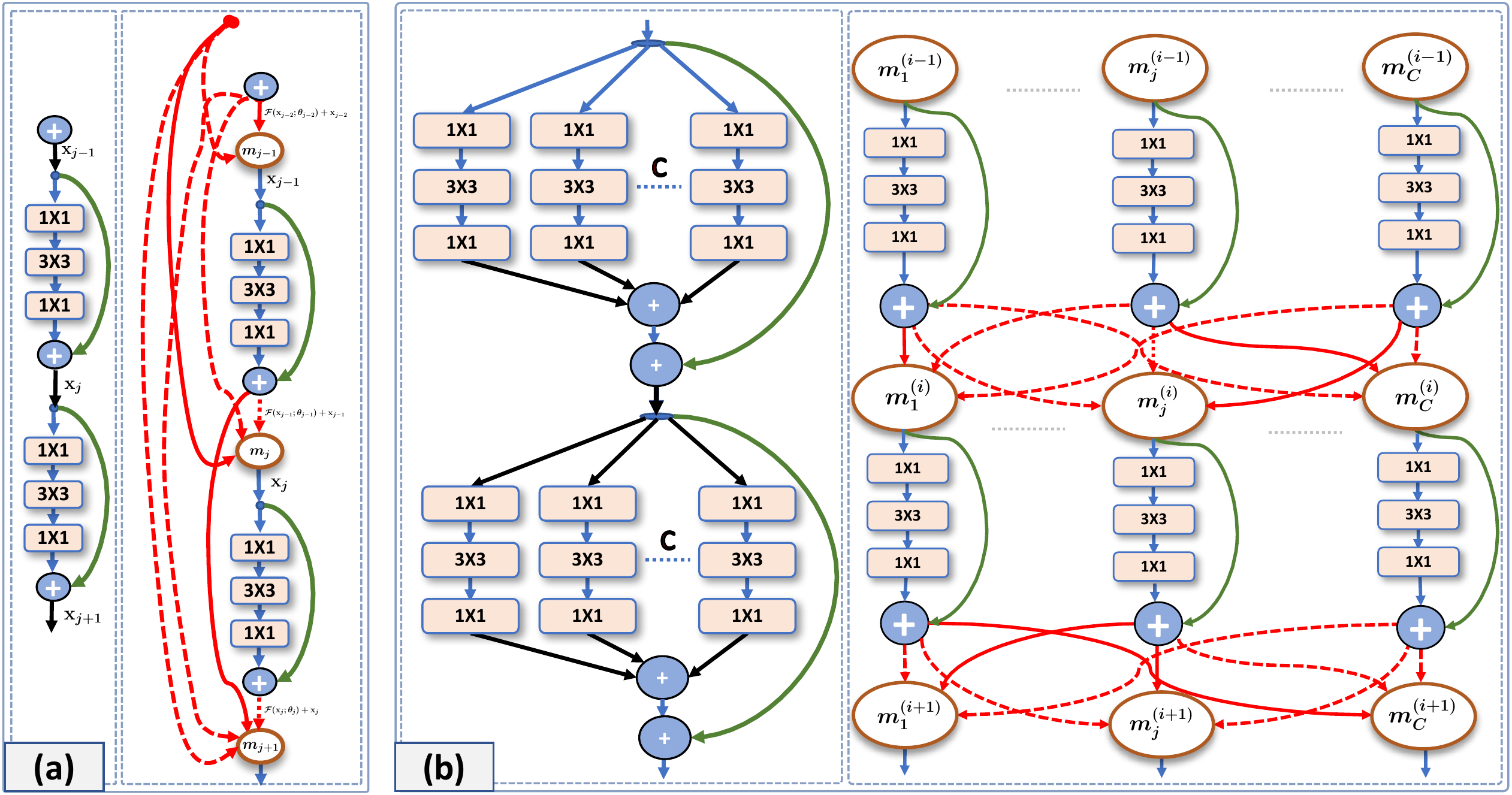}
  \end{center}
  \caption{Application of \maskconnect to two forms of modular network: {\bf(a)} ResNet~\cite{HeEtAl:arXivResNet} and {\bf(b)} multi-branch ResNeXt~\cite{XieGDTH16}. In traditional ResNet (a)(left) the connections between blocks are fixed (black links) so that each block receives input from only the preceding block. Our approach (a)(right) learns the optimal input connections (solid red links) for each individual block from a collection of potential connections (solid and dotted red links). Similarly, in traditional ResNeXt (b)(left) each module consists of $C$ parallel residual blocks which are all aggregated and fed to the next module (black links). \maskconnect (b)(right) replaces the fixed aggregation points of RexNeXt with learnable masks ${\bf m}$ defining the active input connections (solid red links) for each individual residual block.}
  \label{fig:modules}
\end{figure}

The application of our algorithm to ResNets is quite straightforward. ResNets are modular networks obtained by stacking residual blocks. A residual block implements a residual function $\mathcal{F}(.)$ with reference to the layer input. Figure~\ref{fig:modules}(a)(left) illustrates an example of these modular components where the 3 layers in the block implement the residual function $\mathcal{F}({\bf x}; \theta)$. A shortcut connections adds the residual block output $\mathcal{F}({\bf x})$ to its input ${\bf x}$. Thus the complete function $\mathcal{G}(.)$ implemented by a residual block computes $\mathcal{G}({\bf x}; \theta) = \mathcal{F}({\bf x}; \theta) + {\bf x}$. The ResNets originally introduced in~\cite{he2016deep} use a hand-defined connectivity that passes the output of a block to the immediately subsequent block, i.e., ${\bf x}_{j+1} \leftarrow \mathcal{F}({\bf x}_j; \theta_j) + {\bf x}_j$. Here we propose to use \maskconnect to learn the input connections for each individual residual block in the network. This changes the input provided to block $j+1$ in the network to be ${\small {\bf x}_{j+1} \leftarrow \sum_{k=1}^{j} m_{j+1,k} \left[\mathcal{F}({\bf x}_k; \theta_k) + {\bf x}_k\right]}$, where binary parameters $m_{j+1,k}$ are learned automatically by our approach simultaneously with the weights $\theta$ subject to the constraint that $\sum_{k=1}^{j} m_{j+1,k} = K$. This implies that under our model each residual block now receives input from exactly $K$ out of the preceding blocks. The output tensors from the $K$ selected blocks are aggregated using element-wise addition and passed as input to the module. Our experiments present results for varying values of fan-in hyperparameter $K$, which controls the density of connectivity.


\subsection{\maskconnect applied to multi-branch ResNeXt}

The adaptation of \maskconnect to ResNeXt architectures is slightly more complex, as ResNeXt is based on a multi-branch topology.  ResNeXt was motivated by the observation that it is beneficial to arrange residual blocks not only along the depth dimension but also to implement parallel multiple threads of computation feeding from the same input layer. The outputs of the parallel residual blocks are then summed up together with the original input and passed on to the next module. The resulting multi-branch module is illustrated in Figure~\ref{fig:modules}(b)(left). More formally, let \smash{$\mathcal{F}({\bf x}; \theta^{(i)}_j)$} be the transformation implemented by the $j$-th residual block in module $i$-th of the network, where $j=1,\hdots,C$ and $i=1,\hdots,L$, with $L$ denoting the total number of modules stacked on top of each other to form the complete network. The hyperparameter $C$ is called the cardinality of the module and defines the number of parallel branches within each module. The hyperparameter $L$ controls the total depth of the network. Then, in traditional ResNeXt, the output of the $i$-th module is computed as: 
\begin{equation}
{\bf y}_i = {\bf x}_i + \sum_{j=1}^C \mathcal{F}({\bf x}_i; \theta^{(i)}_j)
\end{equation}
In~\cite{XieGDTH16} it was experimentally shown that increasing the cardinality $C$ is a more effective way of improving accuracy compared to increasing depth or the number of filters. In other words, given a fixed budget of parameters, ResNeXt nets were shown to consistently outperform single-branch ResNets.

However, in an attempt to ease network design, a couple of restrictive limitations were embedded in the architecture of ResNeXt modules: (1) the $C$ parallel feature extractors in each module operate on the same input; (2) the number of active branches is constant at all depth levels of the network. 

\maskconnect allows us to remove these restrictions without adding any significant burden on the process of manual network design, with the exception of a single additional integer hyperparameter ($K$) for the entire network. As in ResNeXt, our proposed architecture consists of a stack of $L$ multi-branch modules, each containing $C$ parallel feature extractors. However, differently from ResNeXt, each branch in a module can take a different input. The input pathway of each branch is controlled by a binary mask vector. Let {\small {${\bf m}^{(i)}_{j} = [{m}^{(i)}_{j,1}, {m}^{(i)}_{j,2}, \hdots, {m}^{(i)}_{j,C}]\smash{^\top} \in \{0, 1\}^C$}} be the binary mask vector defining the {\em active} input connections feeding the $j$-th residual block in module $i$. 
We note that under this model we no longer have fixed aggregation nodes summing up {\em all} outputs computed from a module. Instead, the mask {\small \smash{${\bf m}^{(i)}_{j}$}} now determines {\em selectively} for each block which branches from the previous module will be aggregated to form the input to the next block. Under this new scheme, the parallel branches in a module receive different inputs and as such are likely to yield more diverse features. 

As before, different constraints over {\small \smash{${\bf m}^{(i)}_{j}$}} will give rise to different forms of architecture. By introducing the constraint that {\small \smash{$\sum_k {m}^{(i)}_{j,k} = 1$}} for all blocks $j$, then each residual block will receive input from only one branch (since each {\small \smash{${m}^{(i)}_{j,k}$}} must be either 0 or 1). If instead we set {\small \smash{${m}^{(i)}_{j,k}=1$}} for all blocks $j,k$ in each module $i$, then all connections would be active and we would obtain again the fixed ResNeXt architecture. In our experiments we present results obtained by varying the fan-in hyperparameter $K$ such that $1< K < C$. We also note  that it may be possible for a residual block in the network to become unused, as a result of the optimization over the mask values. Thus, at any point in the network the total number of active parallel threads can be any number smaller than or equal to $C$. This implies that a variable branching factor is learned adaptively for the different depths in the network. 

\section{Experiments}
\label{sec:experiments}
We tested our approach on the task of image categorization using two different examples of modularized architecture: ResNet~\cite{he2016deep} and ResNeXt~\cite{XieGDTH16}. We used the following datasets for our evaluation: {CIFAR-10~\cite{Krizhevsky:TR2009}}, CIFAR-100~\cite{Krizhevsky:TR2009}, Mini-ImageNet~\cite{VinyalsBLKW16}, as well as the full {ImageNet~\cite{ImageNet}}. In this paper we include the results achieved on CIFAR-100 and {ImageNet~\cite{ImageNet}}, while the results for {CIFAR-10~\cite{Krizhevsky:TR2009}} and Mini-ImageNet~\cite{VinyalsBLKW16} (showing consistent improvements up to nearly 4\% over fixed connectivity) can be found in Appendix~\ref{appendixA}.

\subsection{CIFAR-100}
\label{subsec:Cifar100Exps}
CIFAR-100 contains images of size 32x32. It consists of 50,000 training images and 10,000 test images. Each image is labeled as belonging to one of 100 possible classes. 
  
\subsubsection{\sloppy \mbox{CIFAR-100~results~based~on~the~{\em ResNet}~architecture.}}~

\begin{wrapfigure}{r}{0.5\textwidth}
\includegraphics[width=0.5\textwidth]{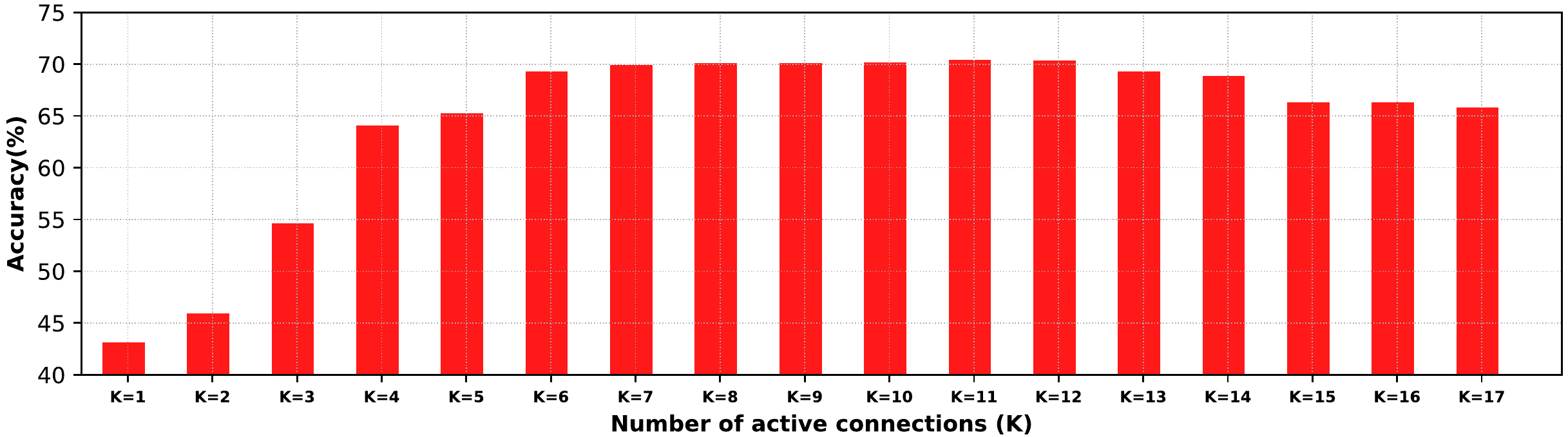}
\captionof{figure}{Varying the fan-in ($K$), i.e., the number of learned active connections to each residual block. The plot reports accuracy achieved by \maskconnect on CIFAR-100 using a {\bf ResNet}-38 architecture ($L=18$ blocks). All models have the same number of parameters (\paramMillion{0.57}). The best accuracy is achieved at $K=10$.}
\label{fig:resnet_cifar100plotK}
\end{wrapfigure}

\paragraph{Effect of fan-in ($K$).} The fan-in hyperparameter ($K$) defines the number of {\em active} input connections feeding each residual block. We study the effect of the fan-in on the performance of models built and trained using our proposed approach. We use residual blocks consisting of two 3x3 convolutional layers. We use a model obtained by stacking $L=18$ residual blocks with total depth of $D=2+2L=38$ layers. We trained and tested this architecture using different fan-in values: $K=1,..,17$. All these models have the same learning capacity as varying $K$ does not affect the number of parameters. The results are shown in Figure~\ref{fig:resnet_cifar100plotK}. We notice that the best accuracy is achieved using $K=10$. Using a very low or very high fan-in yields lower accuracy. However, the algorithm does not appear to be overly sensitive to the fan-in hyperparameter, as a wide range of values for $K$ (from $K=7$ to $K=13$) produce accuracy close to the best.

\begin{table}[t!]

\centering
  \caption{{CIFAR-100} accuracies achieved by models trained using the connectivity of {\bf ResNet}~\cite{HeEtAl:arXivResNet} (Fixed-Prev), a fixed random connectivity (Fixed-Random), and the connectivity learned by our approach (Learned)}
  \label{table:cifar100ResultsResNet}
  {
  \footnotesize
  \begin{tabular}{llc}
    \toprule
{\bf Model}  & \bf Connectivity       &  {\bf Accuracy (\%)}     \\
\midrule
ResNet-38 & Fixed-Prev, K=1~\cite{HeEtAl:arXivResNet}    & 68.54     \\	
		  & Fixed-Random, K=10            		          &   62.67    \\				 
          & \textbf{Learned}, K=10            		  & \textbf{70.40}  \\										 
										 
\midrule

ResNet-74 & Fixed-Prev, K=1~\cite{HeEtAl:arXivResNet}      & 70.64     \\	
		  & Fixed-Random, K=15          		          &       66.93\\				 
          & \textbf{Learned}, K=15             		  & \textbf{72.81}  \\										 
\midrule
          
ResNet-110 & Fixed-Prev, K=1~\cite{HeEtAl:arXivResNet}     & 71.21     \\	
		  & Fixed-Random, K=20          		          &   67.22  \\				 
          & \textbf{Learned}, K=20            		  & \textbf{73.15}  \\										 
\bottomrule
\end{tabular}
}
\end{table}

\paragraph{Varying the model.}  We trained several ResNet models differing in depth, using both \maskconnect as as well as the traditional predefined connectivity. For these experiments we use a stack of $L$ residual blocks with two 3x3 convolutional layers for each block. We choose $L \in \{18, 36, 54\}$ to build networks with depths $D=2+2L$ equal to 38, 74, and 110 layers, respectively. We show the classification accuracy achieved by different models in Table~\ref{table:cifar100ResultsResNet}. We report the results achieved using \maskconnect with fan-in $K=10$, $K=15$, $K=20$ for models of depth $D=38$, $D=74$, $D=110$, respectively. Fixed-Prev denotes the performance of ResNet, where each block is connected to only the previous block ($K=1$). We also include the accuracy achieved by choosing a random connectivity (Fixed-Random) using the same fan-in values $K$ as our approach and training the parameters while keeping the random connectivity fixed. This baseline is useful to show that our model achieves higher accuracy over traditional ResNet not because of the higher number of connections (i.e., $K>1$), but rather because it learns the connectivity. Indeed, the results in Table~\ref{table:cifar100ResultsResNet} show that learning the connectivity using \maskconnect  yields consistently higher accuracy than using multiple random connections or a single connection to the previous block.

\subsubsection{\sloppy CIFAR-100~results~based~on~multi-branch~{\em ResNeXt}.}
\paragraph{Effect of fan-in ($K$).} Even for ResNeXt, we start by studying the effect of the fan-in hyperparameter ($K$). For this experiment we use a model obtained by stacking $L=6$ multi-branch residual modules, each having cardinality $C=8$ (number of branches in each module). We use residual blocks consisting of 3 convolutional layers with a bottleneck implementing dimensionality reduction on the number of feature channels, as shown in Figure~\ref{fig:modules}(b). The bottleneck for this experiment was set to $w=4$. Since each residual block consists of 3 layers, the total depth of the network in terms of learnable layers is $D=2+3L=20$. 

\begingroup
\setlength{\thickmuskip}{0mu}

\begin{figure}[h]
\centering
\parbox{5cm}{
\includegraphics[width=5cm]{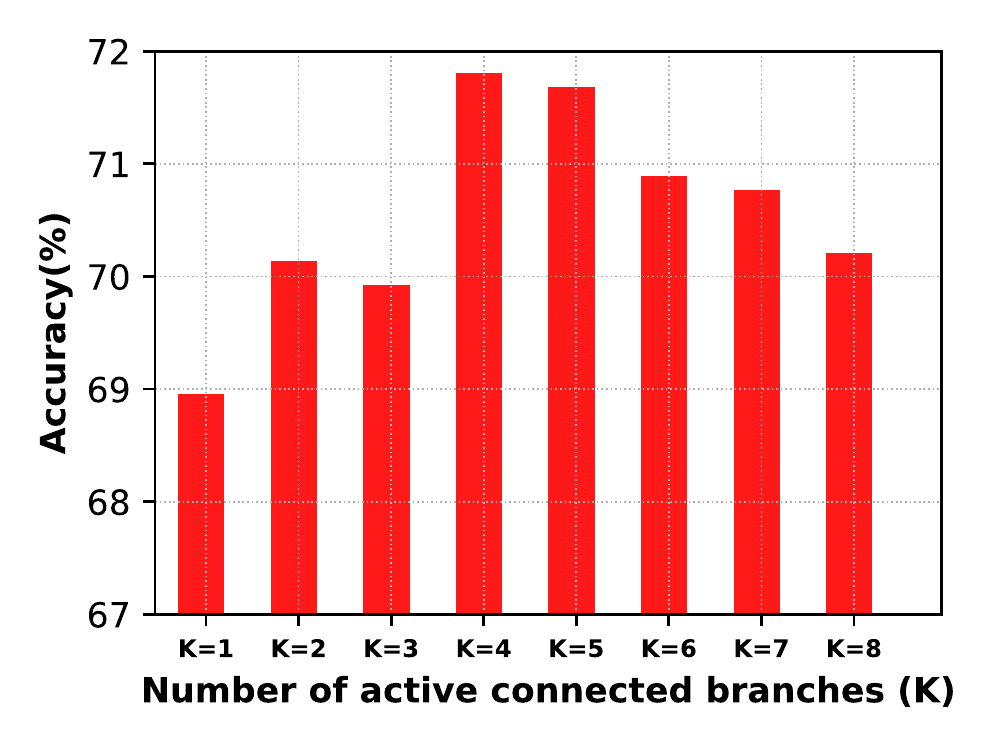}
\caption{Varying the fan-in ($K$) of our model, i.e., the number of active input branches to each residual block. The plot reports accuracy achieved on CIFAR-100 using a network stack of $L=6$ \textbf{ResNeXt} modules having cardinality $C=8$ and bottleneck width $w=4$. All models have the same number of parameters (\paramMillion{0.28}).}
\label{fig:cifar100plotK}}
\qquad
\begin{minipage}{6cm}
\includegraphics[width=6cm]{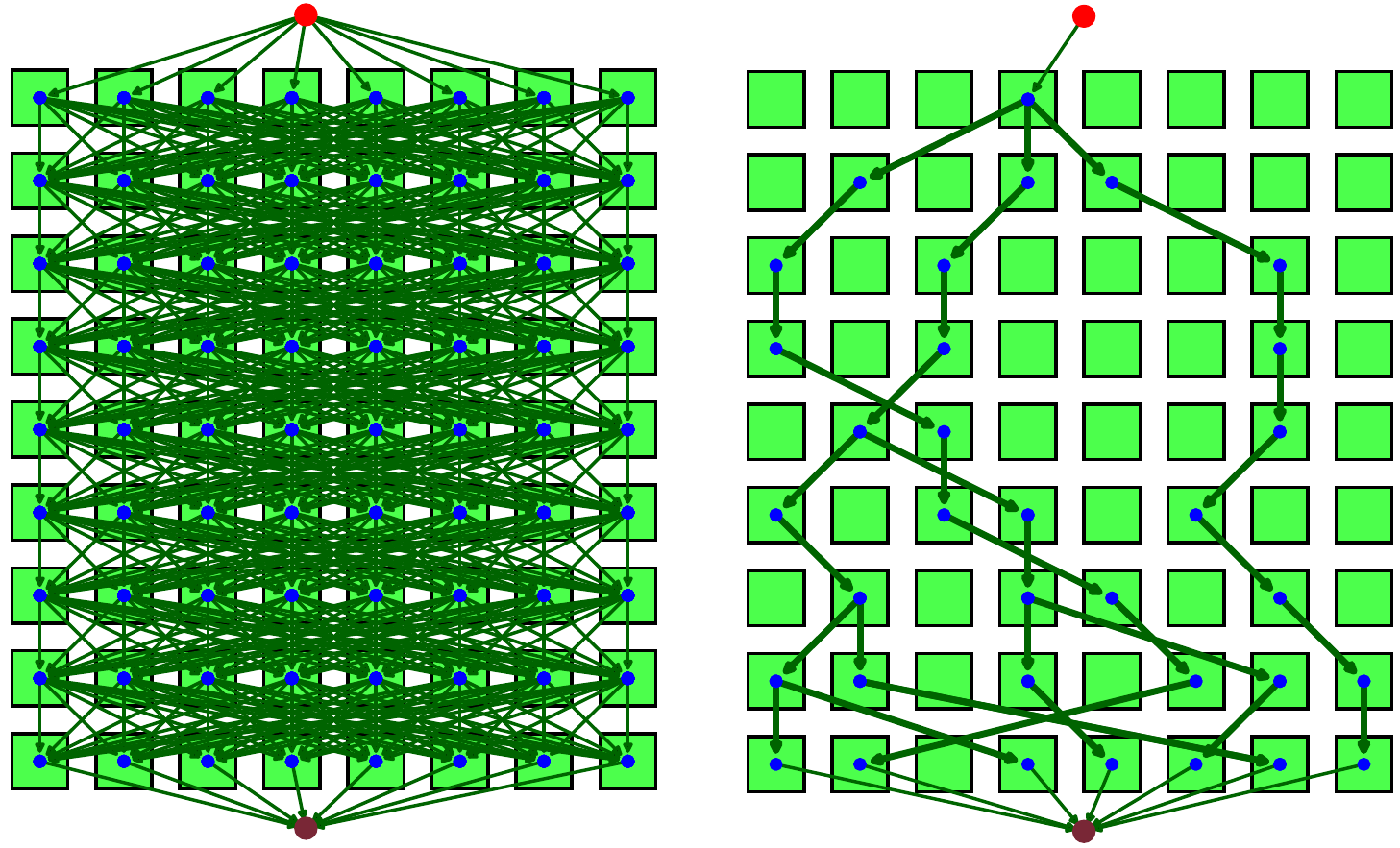}
\caption{A visualization of the fixed connectivity of \textbf{ResNext} (left) vs the connectivity learned by our method (right) using $K=1$. Each green square is a residual block, each row of $C=8$ square is a multi-branch module. 
Arrows indicate pathways connecting residual blocks of adjacent modules. 
It can be noticed that \maskconnect learns sparse connections. The squares without in/out edges are those pruned at the end of learning. This gives rise to a branching factor that varies along the depth of the net.
}
\label{fig:conn}
\end{minipage}
\end{figure}
\endgroup

We trained and tested this architecture using different fan-in values: $K=1,..,8$. Again, varying $K$ does not alter the number of parameters. The results are shown in Figure~\ref{fig:cifar100plotK}. We can see that the best accuracy is achieved by connecting each residual block to $K=4$ branches out of the total $C=8$ in each module. Note that when setting $K=C$, there is no need to learn the masks. In this case each mask is simply replaced by an element-wise addition of the outputs from all the branches. This renders the model equivalent to ResNeXt~\cite{XieGDTH16}, which has fixed connectivity. Based on the results of Figure~\ref{fig:cifar100plotK}, in all our experiments below we use $K=4$ (since it gives the best accuracy) but also $K=1$ since it gives high sparsity which, as we will see shortly, implies savings in number of parameters. 

\paragraph{Varying the models.} In Table~\ref{table:cifar100Results} we show the classification accuracy achieved with ResNeXt models of different depth and cardinality (the details of each model are listed in Appendix~\ref{appendixA}). For each architecture we also include the accuracy achieved with full (as opposed to learned) connectivity, which corresponds to ResNeXt. These results show that learning the connectivity produces consistently higher accuracy than using fixed connectivity, with accuracy gains of up to $2.2\%$ compared to the state-of-the-art ResNeXt model. Furthermore, we can notice that the accuracy of models based on random connectivity (Fixed-Random) is considerably lower compared to our approach, despite having the same connectivity density ($K=4$). This shows that the improvements of our approach over ResNeXt are not due to sparser connectivity but they are rather due to {\em learned} connectivity. We note that these improvements in accuracy come at little computational training cost: the average training time overhead for learning masks and weights is about $39\%$ using our unoptimized implementation compared to learning only the weights given a fixed connectivity. 
    
\paragraph{Parameter savings.} Our proposed approach provides the benefit of automatically identifying residual blocks that are unnecessary. At the end of the training, the unused residual blocks can be pruned away. This yields savings in the number of parameters to store and in test-time computation. In Table~\ref{table:cifar100Results}, columns {\em Train} and {\em Test} under {\em Params} show the original number of parameters (used during training) and the number of parameters after pruning (used at test-time).  Note that for the biggest architecture, our approach using $K=1$ yields a parameter saving of 40\% compared to ResNeXt with full connectivity ($\paramMillion{20.5}$ vs $\paramMillion{34.4}$), while achieving the same accuracy. Thus, in summary, using fan-in $K=4$ gives models that have the same number of parameters as ResNeXt but they yield higher accuracy; using fan-in $K=1$ gives a significant saving in number of parameters and accuracy on par with ResNeXt.

\paragraph{Visualization of the learned connectivity.} Figure~\ref{fig:conn} provides an illustration of  the connectivity learned by \maskconnect for $K=1$ versus the fixed connectivity of ResNeXt for model $\{D=29,w=8,C=8\}$. While ResNeXt feeds the same input to all blocks of a module, our algorithm learns different input pathways for each block and yields a branching factor that varies along depth.
 
\begingroup
\setlength{\thickmuskip}{0mu}

\begin{table}[t!]
  \caption{{CIFAR-100} accuracies achieved by two {\em{\bf ResNeXt}} architectures trained using predefined full connectivity (Fixed-Full)~\cite{XieGDTH16}, random connectivity (Fixed-Random, $K=4$), and the connectivity learned by our algorithm (Learned, $K=1$, $K=4$). Each model was trained 4 times, using different random initializations. We report the best test performance as well as the mean test performance computed from the 4 runs. We list the number of parameters used during training (Params-Train) and the number of parameters obtained after pruning the unused blocks (Params-Test). Our learned connectivity using $K=4$ produces accuracy gains of up to 2.2\% compared to the strong ResNeXt model, while using $K=1$ yields results equivalent to ResNeXt but it induces a significant reduction in number of parameters at test time (e.g., a saving of 40\% for model \{29,64,8\})}
  \label{table:cifar100Results}
  \centering
  {\footnotesize
  \bgroup
\setlength\tabcolsep{0.5em}
  \begin{tabular}{llccccc}
    \toprule
{\bf Architecture}  & \bf Connectivity & \multicolumn{2}{c}{\bf Params}        &  {\bf Accuracy (\%)}     \\
 \cmidrule(lr){3-4}       \cmidrule(lr){5-5} 
\parbox[t]{3.5cm}{\scriptsize \{Depth ($D$), Bottleneck width ($w$), Cardinality ($C$)\}} &   & \emph{\scriptsize Train}  & \emph{\scriptsize Test}   & \emph{\scriptsize best (mean$\pm$std)}  \\ 

\midrule
\multirow{3}{*}{\{29,8,8\}} & Fixed-Full, K=8~\cite{XieGDTH16} & \paramMillion{0.86} & \paramMillion{0.86}       & 73.52 (73.37$\pm$0.13)         \\

										 & \textbf{Learned}, K=1            & \paramMillion{0.86} & \paramMillion{0.65}					 & \textbf{73.91} (73.76$\pm$0.14)           \\
										 
										 & \textbf{Learned}, K=4            & \paramMillion{0.86} &  \paramMillion{0.81}                   & \textbf{75.89} (75.77$\pm$0.12)      \\
										 
										 & Fixed-Random, K=4           & \paramMillion{0.86} &  \paramMillion{0.85}                     & 72.85 (72.66$\pm$0.24)     \\

\midrule
\multirow{3}{*}{\{29,64,8\}} & Fixed-Full, K=8~\cite{XieGDTH16} & \paramMillion{34.4} & \paramMillion{34.4}         & 82.23 (82.12$\pm$0.12)        \\

										    & \textbf{Learned}, K=1            & \paramMillion{34.4} & \paramMillion{20.5}					    & \textbf{82.31} (82.15$\pm$0.15)            \\
										    & \textbf{Learned}, K=4            & \paramMillion{34.4} &   \paramMillion{32.1}                   & \textbf{84.05} (83.94$\pm$0.11)     \\
										    	& Fixed-Random, K=4           & \paramMillion{34.4} &   \paramMillion{34.3}                     & 81.96 (81.73$\pm$0.20)      \\
\bottomrule
\end{tabular}
\egroup
}
\end{table}
\endgroup

\begingroup

\setlength{\thickmuskip}{0mu}

\subsection{ImageNet}
\label{subsec:ImageNetExps}
Finally, we evaluate our approach on the large-scale ImageNet 2012 dataset~\cite{ImageNet}, which includes images of 1000 classes. We train our approach on the training set (1.28M images) and evaluate it on the validation set (50K images). 

\subsubsection{ImageNet results based on the {\em ResNet} architecture.}
For this experiment we use a stack of $L=16$ residual blocks with 3 convolutional layers with a bottleneck architecture. Thus, the total number of layers is $D=2+3L=50$. Compared to the traditional ResNet using fixed connectivity, the same network trained using \maskconnect with fan-in $K=10$ yields a top-1 accuracy gain of $1.94\%$ ($78.09\%$ vs $76.15\%$).

\subsubsection{ImageNet results based on multi-branch {\em ResNeXt}.}
In Table~\ref{table:imagenetResults}, we report the top accuracies for three different ResNeXt architectures. For these experiments we set $K=C/2$. We can observe that for all three architectures, our learned connectivity yields an improvement in accuracy over fixed full connectivity~\cite{XieGDTH16}.

{\footnotesize
\begin{table}[t]
  \caption{ImageNet accuracies (single crop) achieved by different architectures using the predefined connectivity of {\bf ResNeXt} (Fixed-Full) versus the connectivity learned by our algorithm (Learned)} 
  \label{table:imagenetResults}
  \centering
  {\footnotesize
  \begin{tabular}{llcccc}
    \toprule
\bf Architecture  & \bf Connectivity &  \multicolumn{2}{c}{\bf  Accuracy}    \\
\cmidrule(lr){3-4}
\parbox[t]{4.5cm}{\scriptsize \{Depth ($D$), Bottleneck width ($w$), Cardinality ($C$)\}} &   &    \emph{\scriptsize Top-1}  & \emph{\scriptsize Top-5}         \\
    
\midrule
\multirow{2}{*}{\{50,4,32\}} & Fixed-Full, K=32~\cite{XieGDTH16} &    77.8  & 93.3      \\
										 & \textbf{Learned}, K=16            &   \textbf{{79.1}}    &  \textbf{{94.1}}   \\
     
\midrule

\multirow{2}{*}{\{101,4,32\}} & Fixed-Full, K=32~\cite{XieGDTH16} &    78.8  & 94.1        \\
										 &  \textbf{Learned}, K=16            & \textbf{79.5}  &    \textbf{94.5}\\
     
\midrule

\multirow{2}{*}{\{101,4,64\}} & Fixed-Full, K=64~\cite{XieGDTH16} &    79.6   & 94.7     \\
										 &  \textbf{Learned}, K=32            &   \textbf{79.8}  & \textbf{94.8}  \\
     	
\bottomrule
\end{tabular}
}
\end{table}
}
\endgroup

\section{Conclusions}
In this paper we introduced an algorithm to learn the connectivity of deep modular networks. The problem is formulated as a single joint optimization over the weights and connections between modules in the model. We tested our approach on challenging image categorization benchmarks where it led to significant accuracy improvements over the state-of-the-art ResNet and ResNeXt models using fixed connectivity. An added benefit of our approach is that it can automatically identify superfluous blocks, which can be pruned after training without impact on accuracy for more efficient testing and for reducing the number of parameters to store.

While our experiments were carried out on two particular architectures (ResNet and ResNeXt) and a specific form of building block (residual block), we expect the benefits of our approach to extend to other modules and network structures. For example, it could be applied to learn the connectivity of skip-connections in DenseNets~\cite{DenseNets}, which are currently based on predefined connectivity rules. In this paper, our masks perform non-parametric additive aggregation of the branch outputs. It would be interesting to experiment with learnable (parametric) aggregations of the outputs from the individual branches. Our approach is limited to learning connectivity within a given, fixed architecture. Future work will explore the use of learnable masks for full architecture discovery.

\subsubsection{Acknowledgements.} {This work was funded in part by NSF award CNS-120552. We gratefully acknowledge NVIDIA and Facebook for the donation of GPUs used for portions of this work.}


\newpage

\clearpage

\bibliographystyle{splncs04}
\bibliography{egbib}

\appendix

\section{Appendix}\label{appendixA}


\subsection{Pseudocode of \maskconnect applied to ResNet }
\vspace{-0.6cm}
\begin{algorithm}[H]
   \caption{\maskconnect training algorithm For ResNet. }
\label{alg:trainalg1}
\begin{algorithmic}
{\scriptsize
   \STATE {\bfseries Input:} a labeled example $({\bf x}_1, {\bf z}_1)$, $K$: fan-in (number of active inputs out of the preceding blocks), $\eta$: learning rate,  $\pazocal{\ell}$: the loss over the minibatch, $\mathcal{F}({\bf x}; \theta)$ is the residual function, $\tilde{{\bf m}}_{j} \in \left[0,1\right]^j$: real-valued branch masks for block $j$.
   \STATE {\bfseries Output:} updated $\tilde{{\bf m}}_{j}$
 
   \STATE {\bfseries 1. Forward Propagation:}
   \STATE Normalize the real-valued mask to sum up to 1:  $\tilde{{m}}_{j,k}  \leftarrow \frac{\tilde{{m}}_{j,k}}{\sum_{k'=1}^{j} \tilde{{m}}_{j,k'}}$
   \STATE Reset binary mask: ${\bf m}_{j} \leftarrow {\bf 0}$   
   \STATE Draw $K$ {\em distinct} samples from multinomial mask distribution: 
   \STATE $ a_1,a_2,\hdots,a_K \leftarrow  \text{Mult}(\tilde{m}_{j,1},\tilde{m}_{j,2},\hdots,\tilde{m}_{j,j-1})$ 
   \STATE Set active binary mask based on drawn samples: \\$m_{j,a_k} \leftarrow 1 \text{ for } k=1,...,K$
   \STATE Compute input ${\bf x}_{j}$ to $j$-th block ($j\geq2$):
    ${\small {\bf x}_{j} \leftarrow \sum_{k=1}^{j-1} m_{j,k} \left[\mathcal{F}({\bf x}_k; \theta_k) + {\bf x}_k\right]}$
  
   \STATE {\bfseries 2. Backward Propagation:}  
   \STATE Compute $\frac{\partial \pazocal{\ell}}{\partial {\bf x}_{j}}$ 
   \STATE Compute $\frac{\partial \pazocal{\ell}}{\partial {\bf x}_{k}}$ from $\frac{\partial \pazocal{\ell}}{\partial {\bf x}_{j}}$ and $m_{j,k}$ for $j>k$
   
   \STATE {\bfseries 3. Parameter Update:}   
   \STATE Compute $\frac{\partial \pazocal{\ell}}{\partial m_{j,k}}$  given  $\frac{\partial \pazocal{\ell}}{\partial {\bf x}_{j}}$  and $\left[\mathcal{F}({\bf x}_k; \theta_k) + {\bf x}_k\right]$
 
\STATE $\tilde{m}_{j,k} \leftarrow $ \text{clip}($\tilde{m}_{j,k} - \eta \cdot \frac{\partial \pazocal{\ell}}{\partial m_{j,k} } )$
 }
\end{algorithmic}
\end{algorithm}

\vspace{-1.2cm}


\subsection{Pseudocode of \maskconnect applied to multi-branch ResNeXt}
\vspace{-0.6cm}
\begin{algorithm}[H]
   \caption{\maskconnect training algorithm For ResNeXt. }
\label{alg:trainalg2}
\begin{algorithmic}
{\scriptsize
   \STATE {\bfseries Input:} a labeled examples $({\bf x}, {\bf z})$, $C$: cardinality (number of  branches), $K$: fan-in (number of active branch connections), $\eta$: learning rate,  $\pazocal{\ell}$: the loss over the minibatch, $\tilde{{\bf m}}^{(i)}_j \in \left[0,1\right]^C$: real-valued branch masks for block $j$ in module $i$ from previous training iteration.
   \STATE {\bfseries Output:} updated $\tilde{{\bf m}}^{(i)}_j$
 
   \STATE {\bfseries 1. Forward Propagation:}
   \STATE Normalize the real-valued mask to sum up to 1:  $\tilde{{m}}^{(i)}_{j,k}  \leftarrow \frac{\tilde{{m}}^{(i)}_{j,k}}{\sum_{k'=1}^{C} \tilde{{m}}^{(i)}_{j,k'}},$ for $j=1,\hdots, C$
   \STATE Reset binary mask: ${\bf m}^{(i)}_j \leftarrow {\bf 0}$   
   \STATE Draw $K$ {\em distinct} samples from multinomial mask distribution: 
   \STATE $ a_1,a_2,\hdots,a_K \leftarrow  \text{Mult}(\tilde{m}^{(i)}_{j,1},\tilde{m}^{(i)}_{j,2},\hdots,\tilde{m}^{(i)}_{j,C})$ 
   \STATE Set active binary mask based on drawn samples: \\$m^{(i)}_{j,a_k} \leftarrow 1 \text{ for } k=1,...,K$
   \STATE Compute input ${\bf x}^{(i)}_j$ of block $j$ in module $i$ given branch activations ${\bf y}^{(i-1)}_k$ and mask ${\bf m}^{(i)}_j$:
    ${\bf x}^{(i)}_j \leftarrow \sum_{k=1}^{C} {m}^{(i)}_{j,k} \cdot {\bf y}^{(i-1)}_k$
  
   \STATE {\bfseries 2. Backward Propagation:}  
   \STATE Compute $\frac{\partial \pazocal{\ell}}{\partial {\bf x}^{(i)}_{j}}$ from $\frac{\partial \pazocal{\ell}}{\partial {\bf y}^{(i)}_{j}}$
   \STATE Compute $\frac{\partial \pazocal{\ell}}{\partial {\bf y}^{(i-1)}_{k}}$ from $\frac{\partial \pazocal{\ell}}{\partial {\bf x}^{(i)}_{j}}$ and $m^{(i)}_{j,k}$
   
   \STATE {\bfseries 3. Parameter Update:}   
   \STATE Compute $\frac{\partial \pazocal{\ell}}{\partial m^{(i)}_{j,k}}$  given  $\frac{\partial \pazocal{\ell}}{\partial {\bf x}^{(i)}_{j}}$  and ${\bf y}^{(i-1)}_{k}$
 
\STATE $\tilde{m}^{(i)}_{j,k} \leftarrow $ \text{clip}($\tilde{m}^{(i)}_{j,k} - \eta \cdot \frac{\partial \pazocal{\ell}}{\partial m^{(i)}_{j,k} } )$
 }
\end{algorithmic}
\end{algorithm}



\subsection{Experiments on CIFAR-10 based on ResNeXt Architecture}
\label{sec:Cifar10Exps}

The {CIFAR-10} dataset consists of color images of size 32x32. The training set contains 50,000 images, the testing set 10,000 images. Each image {in CIFAR-10} is categorized into one of 10 possible classes. In Table~\ref{table:cifar10Results}, we report the performance of different models trained on {CIFAR-10}. From these results we can observe that our models using learned connectivity achieve consistently better performance over the equivalent models trained with the fixed connectivity~\cite{XieGDTH16}. 

\begin{table}[ht]
  \caption{{CIFAR-10} accuracies {(single crop)} achieved by different multi-branch architectures trained using the predefined connectivity of ResNeXt (Fixed-Full) versus the connectivity learned by our algorithm (Learned). Each model was trained 4 times, using different random initializations. For each model we report the best test performance as well as the mean test performance computed from the 4 runs} 
  \label{table:cifar10Results}
  \centering
  {\small
  \begin{tabular}{llc}
    \toprule
    
{\bf Architecture}  & \bf Connectivity &  {\bf  Accuracy (\%)}   \\

 \cmidrule(lr){3-3}    
\parbox[t]{3.0cm}{\{\scriptsize Depth ($D$), Bottleneck width ($w$), Cardinality ($C$)\}} &    &  \multicolumn{1}{p{3.1cm}}{\centering \emph{Top-1} \\  \centering best (mean$\pm$std) } \\ 


\midrule
\multirow{3}{*}{\{20,4,8\}} & Fixed-Full K=8~\cite{XieGDTH16} & 91.39 (91.13$\pm$0.11)        \\
										 & \textbf{Learned} K=4        &    \textbf{92.85}  (92.76$\pm$0.10)   \\

\midrule
\multirow{3}{*}{\{29,4,8\}} & Fixed-Full K=8~\cite{XieGDTH16}  & 92.77 (92.65$\pm$0.09)          \\
										 & \textbf{Learned} K=4             &     \textbf{93.88}  (93.76$\pm$0.12)    \\

\midrule
\multirow{3}{*}{\{29,8,8\}} & Fixed-Full K=8~\cite{XieGDTH16}  &93.26 (93.14$\pm$0.11)          \\
										 & \textbf{Learned} K=4            &    \textbf{95.11}  (94.96$\pm$0.12)  \\

\midrule
\multirow{3}{*}{\{29,64,8\}} & Fixed-Full K=8~\cite{XieGDTH16}  & 96.35 (96.23$\pm$0.12)      \\
										    & \textbf{Learned} K=4    &  \textbf{96.83}  (96.73$\pm$0.11)  \\

\bottomrule
\end{tabular}
}
\end{table}

\begin{table}[h]
  \caption{Mini-ImageNet accuracies achieved by different multi-branch networks trained using the predefined full connectivity of ResNeXt (Fixed-Full) versus the connectivity learned by our algorithm (Learned). Additionally, we include models trained using random fixed connectivity (Fixed-Random) for $K=4$. For each model we report the best and the mean test performance computed from 4 different training runs. Our method for joint learning of weights and connectivity yields a gain of over $3\%$ in Top-1 accuracy over ResNeXt, which uses the same architectures but a fixed branch connectivity} 
%
%
%
  \label{table:miniimagenetResults}
  \centering
  {\small
  \begin{tabular}{llc}
    \toprule
\bf Architecture  & \bf Connectivity &  {\bf  Accuracy}    \\
 \cmidrule(lr){3-3}  
\parbox[t]{3.0cm}{\{\scriptsize Depth ($D$), Bottleneck width ($w$), Cardinality ($C$\}} &    &\multicolumn{1}{p{3.1cm}}{\centering \emph{Top-1} \\  \centering best (mean$\pm$std) } \\ 

\midrule
\multirow{3}{*}{\{20,4,8\}} & Fixed-Full K=8~\cite{XieGDTH16}        & 62.12 (61.86$\pm$0.15)          \\
										 & \textbf{Learned} K=4                        &  \textbf{66.09} (65.94$\pm$0.16)    \\
     								    & Fixed-Random K=4                       &  62.42 (61.81$\pm$0.32)           \\

\midrule
\multirow{2}{*}{\{29,8,8\}} & Fixed-Full K=8~\cite{XieGDTH16}  & 68.11 (67.89$\pm$0.19)            \\
										 & \textbf{Learned} K=4                    &     \textbf{71.36}  (71.18$\pm$0.19)   \\
										 & Fixed-Random K=4                     &  67.97 (67.53$\pm$0.20)      \\

\bottomrule
\end{tabular}
}
\end{table}

\subsection{Experiments on Mini-ImageNet based on ResNeXt Architecture}
\label{subsec:MiniExps}
Mini-ImageNet  is a subset of the full ImageNet~\cite{ImageNet} dataset. It was used in~\cite{VinyalsBLKW16,ravi2017optimization}. It is created by randomly selecting 100 classes from the full ImageNet~\cite{ImageNet}. For each class, 600 images are randomly selected. We use 500 examples per class for training, and the other 100 examples per class for testing. The selected images are resized to size 84x84 pixels as in~\cite{VinyalsBLKW16,ravi2017optimization}. The advantage of this dataset is that it poses the recognition challenges typical of the ImageNet photos but at the same time it does not need require the powerful resources needed to train on the full ImageNet dataset. This allows to include the additional baselines involving random fixed connectivity (Fixed-Random).

We report the performance of different models trained on Mini-ImageNet in Table~\ref{table:miniimagenetResults}. From these results, we see that our models using learned connectivity with fan-in $K$=4 yield a nice accuracy gain over the same models trained with the fixed full connectivity of ResNeXt~\cite{XieGDTH16}. The absolute improvement (in Top-1 accuracy) is 3.87\% for the 20-layer network and 3.17\% for the 29-layer network. We can notice that the accuracy of the models with fixed random connectivity (Fixed-Random) is considerably lower compared to our nets with learned connectivity, despite having the same connectivity density ($K=4$). This shows that the improvement of our approach over ResNeXt is not due to sparser connectivity but it is rather due to {\em learned} connectivity.



\begin{table}[!t]
  \caption{Specifications of the architectures used in our experiments on the CIFAR-10 and CIFAR-100 datasets based on ResNeXt architecture. The architectures differ in terms of depth ($D$), bottleneck width ($w$), and cardinality ($C$).  Inside the brackets we specify the residual block used in each multi-branch module by listing the number of input channels, the size of the convolutional filters, as well as the number of filters (number of output channels). To the right of each bracket we list the cardinality (i.e., the number of parallel branches in the module). $\times 2$ means that the same multi-branch module is stacked twice. The first layer for all models is a convolutional layer with 16 filters of size $3 \times 3$. The last layer performs global average pooling followed by a softmax} 
  \label{table:cifar100_archs}
  \centering
  {\tiny
  \begin{tabular}{llll}
    \toprule
    
\bf {\{$D$=20, $w$=4, $C$=8\}} & \bf {\{$D$=29, $w$=4, $C$=8\}} & \bf {\{$D$=29, $w$=8, $C$=8\}}  & \bf {\{$D$=29, $w$=64, $C$=8\}}       \\
\midrule
\midrule


3, 3$\times$3, 16			
&3, 3$\times$3, 16   
&3, 3$\times$3, 16   
&3, 3$\times$3, 64   

\\

\midrule
\\
\Bigg[ \parbox[]{1.2cm}{16, 1$\times$1, 4\\ 4,  3$\times$3, 4\\ 4,  1$\times$1, 64} \Bigg] ($C$=8)	&
\Bigg[ \parbox[]{1.2cm}{16, 1$\times$1, 4\\ 4,  3$\times$3, 4\\ 4,  1$\times$1, 64} \Bigg] ($C$=8) &
\Bigg[ \parbox[]{1.2cm}{16, 1$\times$1, 8\\ 8,  3$\times$3, 8\\ 8,  1$\times$1, 64} \Bigg] ($C$=8) &
\Bigg[ \parbox[]{1.3cm}{64, 1$\times$1, 64\\ 64,  3$\times$3, 64\\ 64,  1$\times$1, 256} \Bigg] ($C$=8)
 \\ \\ 
\Bigg[ \parbox[]{1.2cm}{64, 1$\times$1, 4\\ 4,  3$\times$3, 4\\ 4,  1$\times$1, 64} \Bigg] ($C$=8) &
\Bigg[ \parbox[]{1.2cm}{64, 1$\times$1, 4\\ 4,  3$\times$3, 4\\ 4,  1$\times$1, 64} \Bigg] ($C$=8),$\times$2 &
\Bigg[ \parbox[]{1.2cm}{64, 1$\times$1, 8\\ 8,  3$\times$3, 8\\ 8,  1$\times$1, 64} \Bigg] ($C$=8),$\times$2 &
\Bigg[ \parbox[]{1.3cm}{256, 1$\times$1, 64\\ 64,  3$\times$3, 64\\ 64,  1$\times$1, 256} \Bigg] ($C$=8),$\times$2

\\

\midrule
\Bigg[ \parbox[]{1.5cm}{64, 1$\times$1, 8\\ 8,  3$\times$3, 8\\ 8,  1$\times$1, 128} \Bigg] ($C$=8) &
\Bigg[ \parbox[]{1.5cm}{64, 1$\times$1, 8\\ 8,  3$\times$3, 8\\ 8,  1$\times$1, 128} \Bigg] ($C$=8) &
\Bigg[ \parbox[]{1.5cm}{64, 1$\times$1, 16\\ 16,  3$\times$3, 16\\ 16,  1$\times$1, 128} \Bigg] ($C$=8) &
\Bigg[ \parbox[]{1.5cm}{256, 1$\times$1, 128\\ 128,  3$\times$3, 128\\ 128,  1$\times$1, 512} \Bigg] ($C$=8)   
 \\ \\
\Bigg[ \parbox[]{1.5cm}{128, 1$\times$1, 8\\ 8,  3$\times$3, 8\\ 8,  1$\times$1, 128} \Bigg] ($C$=8)&
\Bigg[ \parbox[]{1.5cm}{128, 1$\times$1, 8\\ 8,  3$\times$3, 8\\ 8,  1$\times$1, 128} \Bigg] ($C$=8),$\times$2 &
\Bigg[ \parbox[]{1.5cm}{128, 1$\times$1, 16\\ 16,  3$\times$3, 16\\ 16,  1$\times$1, 128} \Bigg] ($C$=8),$\times$2  & 
\Bigg[ \parbox[]{1.5cm}{512, 1$\times$1, 128\\ 128,  3$\times$3, 128\\ 128,  1$\times$1, 512} \Bigg] ($C$=8),$\times$2

\\

\midrule

\Bigg[ \parbox[]{1.55cm}{128, 1$\times$1, 16\\ 16,  3$\times$3, 16\\ 16,  1$\times$1, 256} \Bigg] ($C$=8) &
\Bigg[ \parbox[]{1.55cm}{128, 1$\times$1, 16\\ 16,  3$\times$3, 16\\ 16,  1$\times$1, 256} \Bigg] ($C$=8) &
\Bigg[ \parbox[]{1.55cm}{128, 1$\times$1, 32\\ 32,  3$\times$3, 32\\ 32,  1$\times$1, 256} \Bigg] ($C$=8) &
\Bigg[ \parbox[]{1.55cm}{512, 1$\times$1, 256\\ 256,  3$\times$3, 256\\ 256,  1$\times$1, 1024} \Bigg] ($C$=8)

\\ \\
\Bigg[ \parbox[]{1.55cm}{256, 1$\times$1, 16\\ 16,  3$\times$3, 16\\ 16,  1$\times$1, 256} \Bigg] ($C$=8) &
\Bigg[ \parbox[]{1.55cm}{256, 1$\times$1, 16\\ 16,  3$\times$3, 16\\ 16,  1$\times$1, 256} \Bigg] ($C$=8),$\times$2   &
\Bigg[ \parbox[]{1.55cm}{256, 1$\times$1, 32\\ 32,  3$\times$3, 32\\ 32,  1$\times$1, 256} \Bigg] ($C$=8),$\times$2   &
\Bigg[ \parbox[]{1.55cm}{1024, 1$\times$1, 256\\ 256,  3$\times$3, 256\\ 256,  1$\times$1, 1024} \Bigg] ($C$=8),$\times$2    

\\

\midrule
{Average Pool} & {Average Pool} &{Average Pool} & {Average Pool}\\
{100 fc, softmax} & {100 fc, softmax} & {100 fc, softmax} & {100 fc, softmax}\\

\bottomrule

\end{tabular}
}
\end{table}

\subsection{Visualizations of learned connectivity based on ResNeXt Architecture}

\begin{figure}
\includegraphics[width=\linewidth]{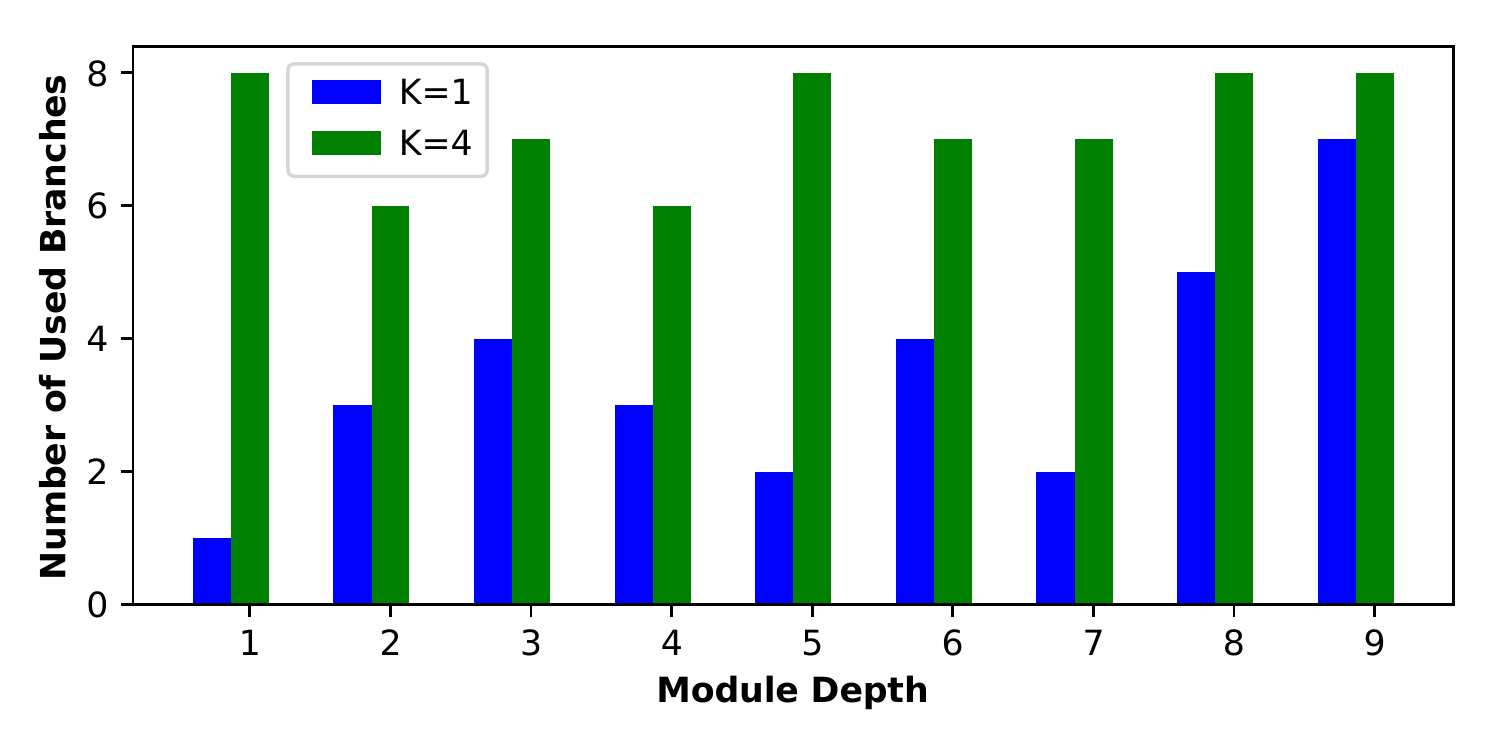}
\caption{Number of active branches as a function of module depth for  model $\{D=29,w=4,C=8\}$ trained on CIFAR-100. We report how the number of active branches varies for model trained with fan-in $K=1$ as well as for the net trained with $K=4$. The setting $K=1$ tends to leave many blocks unused, especially in the early modules of the network.}
\label{fig:branchingCIFAR100}
\end{figure}

TheSupp plot in Figure~\ref{fig:branchingCIFAR100} shows how the number of active branches varies as a function of the module depth for  model $\{D=29,w=4,C=8\}$ trained on CIFAR-100. For $K=1$, we can observe that the number of active branches tends to be larger for deep modules (closer to the output layer) compared to early modules (closer to the input). We observed this phenomenon consistently for all architectures. This suggests that having many parallel threads of computation is particularly important in deep layers of the network. Conversely, the setting $K=4$ tends to produce a fairly uniform number of active branches across the modules and the number is quite close to the maximum value $C$. For this reason, there is little saving in terms of number of parameters when using $K=4$, as there are rarely unused blocks.

\begin{figure}
\includegraphics[width=\linewidth]{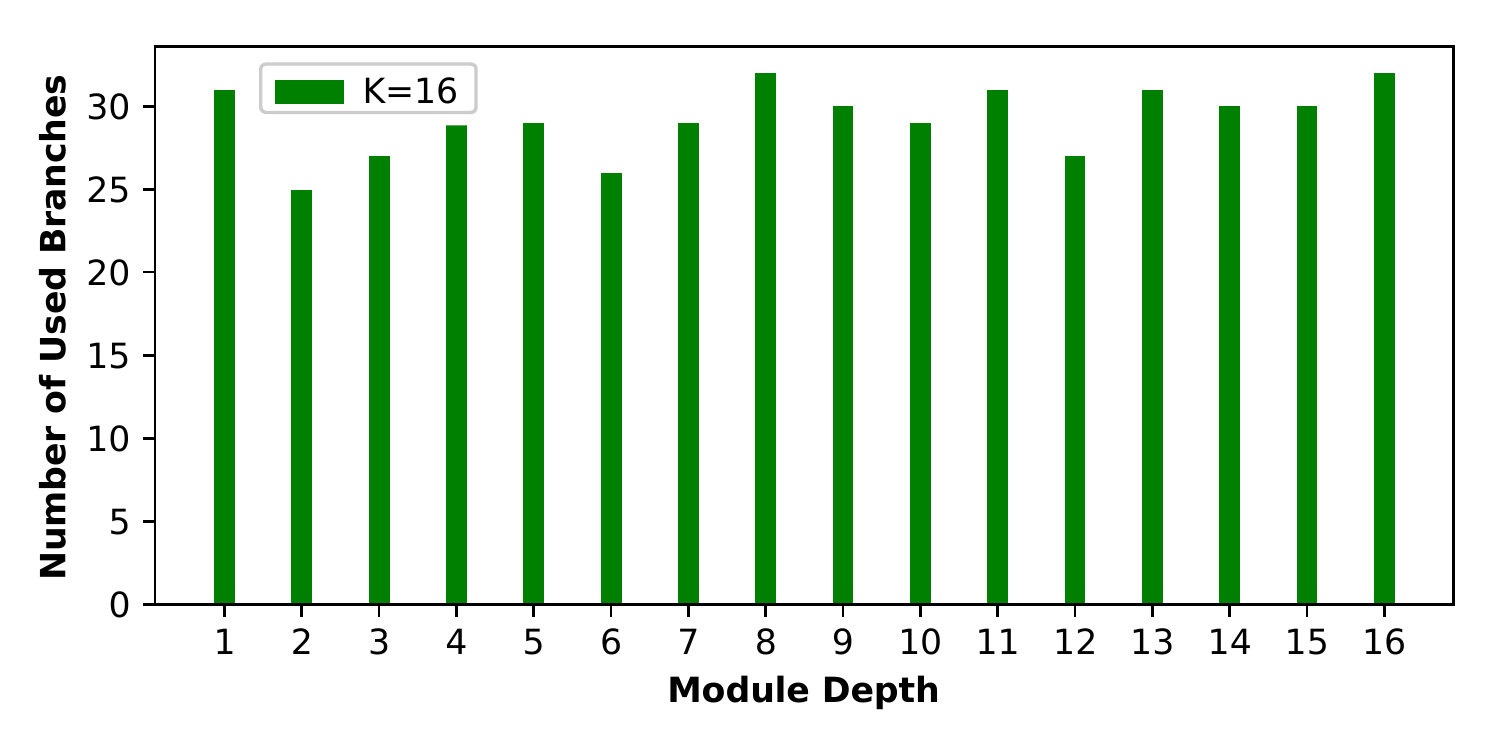}
\caption{Number of active branches as a function of module depth for  model $\{D=50,w=4,C=32\}$ trained on ImageNet, using fan-in $K=16$.}
\label{fig:branchingImageNet}
\end{figure}

The plot in Figure~\ref{fig:branchingImageNet} shows the number of active branches as a function of module depth for model 
$\{D=50,w=4,C=32\}$ trained on ImageNet, using $K=16$.


%
%

\subsection{Implementation details} 

\subsubsection{Architectures and settings for experiments on CIFAR-100 based on ResNet}~

The architectures used in theses experiments are the same used in the original ResNet paper~\cite{he2016deep} for CIFAR-10 experiments except for the last fully-connected layer and softmax which have output size of 100 instead of 10. We used the data augmentation strategy where four pixels are padded on each side of the input image, and a 32x32 crop is randomly sampled from the padded image or its horizontal flip, with per-pixel mean subtracted~\cite{AlexNet}. For testing, we use the original 32x32 image.  The stacks have output feature map of size 32, 16, and 8 respectively. The models are trained on 2 GPUs with a mini-batch size of 128, with a weight decay of 0.0001 and momentum of 0.9. We adopt {\em four} incremental training phases with a total of 80 epochs. In {\em phase 1} we train the model for 30 epochs with a learning rate of 0.1 for the convolutional and fully-connected layers, and a learning rate of 0.3 for the masks. In {\em phase 2} we freeze the connectivity by setting as active connections for each block those corresponding to its top-$K$ values in the masks.  With these fixed learned connectivity, we finetune the model from {\em phase 1} for 30 epochs with a learning rate of 0.1 for the weights. Then, in {\em phase 3} we finetune the weights of the model from {\em phase 2} for 10 epochs with a learning rate of 0.01 using again the fixed learned connectivity from phase 1. Finally, in {\em phase 4} we finetune the weights of the model from {\em phase 3} for 10 epochs with a learning rate of 0.001.

\subsubsection{Architectures and settings for experiments on ImageNet based on ResNet}~

The architectures for our ImageNet experiments are those specified in the original ResNet paper~\cite{he2016deep}. For these experiments, we follow the data augmentation strategy described in~\cite{he2016deep}. The size of the input image is 224x224 and it is randomly cropped from the resized original image. We use a mini-batch size of 256, with a weight decay of 0.0001 and a momentum of 0.9. We use {\em four} incremental training phases with a total of 120 epochs. In {\em phase 1} we train the model for 30 epochs with a learning rate of 0.1 for the convolutional and fully-connected layers, and a learning rate of 0.3 for the masks. In {\em phase 2} we finetune the model from {\em phase 1} for another 30 epochs with a learning rate of 0.1 and a learning rate of 0.0 for the masks (i.e., we use the fixed connectivity learned in phase 1). In {\em phase 3} we finetune the weights from {\em phase 2} for 30 epochs with a  learning rate of 0.01 and the learning rate of the masks is 0.0. Finally, in {\em phase 4} we finetune the weights from {\em phase 3} for 30 epochs with a  learning rate of 0.001 while the learning rate of the masks is still set to 0.0.

\subsubsection{Architectures and settings for experiments on CIFAR-100 and CIFAR-10 based on ResNeXt}~

The specifications of the architectures used in all our experiments on CIFAR-10 and CIFAR-100 are given in Table~\ref{table:cifar100_archs}. 

Several of these architectures are those presented in the original ResNeXt paper~\cite{XieGDTH16} and are trained using the same setup, including the data augmentation strategy. Four pixels are padded on each side of the input image, and a 32x32 crop is randomly sampled from the padded image or its horizontal flip, with per-pixel mean subtracted~\cite{AlexNet}. For testing, we use the original 32x32 image.  The stacks have output feature map of size 32, 16, and 8 respectively. The models are trained on 8 GPUs with a mini-batch size of 128 (16 per GPU), with a weight decay of 0.0005 and momentum of 0.9. We adopt {\em four} incremental training phases with a total of 320 epochs. In {\em phase 1} we train the model for 120 epochs with a learning rate of 0.1 for the convolutional and fully-connected layers, and a learning rate of 0.2 for the masks. In {\em phase 2} we freeze the connectivity by setting as active connections for each block those corresponding to its top-$K$ values in the masks.  With these fixed learned connectivity, we finetune the model from {\em phase 1} for 100 epochs with a learning rate of 0.1 for the weights. Then, in {\em phase 3} we finetune the weights of the model from {\em phase 2} for 50 epochs with a learning rate of 0.01 using again the fixed learned connectivity from phase 1. Finally, in {\em phase 4} we finetune the weights of the model from {\em phase 3} for 50 epochs with a learning rate of 0.001.

\subsubsection{Architectures and settings for experiments on ImageNet based on ResNeXt}~

The architectures for our ImageNet experiments are those specified in the original ResNeXt paper~\cite{XieGDTH16}. 

Also for these experiments, we follow the data augmentation strategy described in~\cite{XieGDTH16}. The input image has size 224x224 and it is randomly cropped from the resized original image. We use a mini-batch size of 256 on 8 GPUs (32 per GPU), with a weight decay of 0.0001 and a momentum of 0.9. We use {\em four} incremental training phases with a total of 120 epochs. In {\em phase 1} we train the model for 30 epochs with a learning rate of 0.1 for the convolutional and fully-connected layers, and a learning rate of 0.2 for the masks. In {\em phase 2} we finetune the model from {\em phase 1} for another 30 epochs with a learning rate of 0.1 and a learning rate of 0.0 for the masks (i.e., we use the fixed connectivity learned in phase 1). In {\em phase 3} we finetune the weights from {\em phase 2} for 30 epochs with a  learning rate of 0.01 and the learning rate of the masks is 0.0. Finally, in {\em phase 4} we finetune the weights from {\em phase 3} for 30 epochs with a  learning rate of 0.001 while the learning rate of the masks is still set to 0.0.

\subsubsection{Architectures and settings for experiments on Mini-ImageNet based on ResNeXt}~

For the experiments on the Mini-ImageNet dataset, a 64x64 crop is randomly sampled from the scaled 84x84 image or its horizontal flip, with per-pixel mean subtracted~{AlexNet}. For testing, we use the center 64x64 crop. The specifications of the models are identical to the CIFAR-100 models used in the previous subsection, except that the first input convolutional layer in the network is followed by a max pooling layer. The models are trained on 8 GPUs with a mini-batch size of 256 (32 per GPU), with a weight decay of 0.0005 and momentum of 0.9. Similar to training CIFAR-100 dataset, we also adopt {\em four} incremental training phases with a total of 320 epochs.

\begin{table}[t]
  \caption{Mini-ImageNet architectures with varying depth ($D$), and bottleneck width ($w$).  Inside the brackets we specify the residual block used in each multi-branch module by listing the number of input channels, the size of the convolutional filters, as well as the number of filters (number of output channels). To the right of each bracket we list the cardinality ($C$) (i.e., the number of parallel branches in the module). $\times 2$ means that the same multi-branch module is stacked twice} 
  \label{table:miniimagenet_archs}
  \centering
  {\tiny
  \begin{tabular}{ll}
    \toprule
    
\bf {\{$D$=20, $w$=4, $C$=8\}}  & \bf {\{$D$=29, $w$=8, $C$=8\}}       \\
\midrule
\midrule


3, 3$\times$3, 16			
& 3, 3$\times$3, 16   

\\
\midrule

{(Max Pool,3$\times$3, stride=2)} & {(Max Pool, 3$\times$3, stride=2)}\\

\midrule
\\
\Bigg[ \parbox[]{1.2cm}{16, 1$\times$1, 4\\ 4,  3$\times$3, 4\\ 4,  1$\times$1, 64} \Bigg] ($C$=8)	&
\Bigg[ \parbox[]{1.2cm}{16, 1$\times$1, 8\\ 8,  3$\times$3, 8\\ 8,  1$\times$1, 64} \Bigg] ($C$=8)
 \\ \\ 
\Bigg[ \parbox[]{1.2cm}{64, 1$\times$1, 4\\ 4,  3$\times$3, 4\\ 4,  1$\times$1, 64} \Bigg] ($C$=8) &
\Bigg[ \parbox[]{1.2cm}{64, 1$\times$1, 8\\ 8,  3$\times$3, 8\\ 8,  1$\times$1, 64} \Bigg] ($C$=8), $\times$2

\\

\midrule
\Bigg[ \parbox[]{1.3cm}{64, 1$\times$1, 8\\ 8,  3$\times$3, 8\\ 8,  1$\times$1, 128} \Bigg] ($C$=8) &
\Bigg[ \parbox[]{1.3cm}{64, 1$\times$1, 16\\ 16,  3$\times$3, 16\\ 16,  1$\times$1, 128} \Bigg] ($C$=8)
 \\ \\
\Bigg[ \parbox[]{1.3cm}{128, 1$\times$1, 8\\ 8,  3$\times$3, 8\\ 8,  1$\times$1, 128} \Bigg] ($C$=8)&
\Bigg[ \parbox[]{1.3cm}{128, 1$\times$1, 16\\ 16,  3$\times$3, 16\\ 16,  1$\times$1, 128} \Bigg] ($C$=8), $\times$2

\\

\\
\\

\midrule

\Bigg[ \parbox[]{1.4cm}{128, 1$\times$1, 16\\ 16,  3$\times$3, 16\\ 16,  1$\times$1, 256} \Bigg] ($C$=8) &
\Bigg[ \parbox[]{1.4cm}{128, 1$\times$1, 32\\ 32,  3$\times$3, 32\\ 32,  1$\times$1, 256} \Bigg] ($C$=8)

\\ \\
\Bigg[ \parbox[]{1.4cm}{256, 1$\times$1, 16\\ 16,  3$\times$3, 16\\ 16,  1$\times$1, 256} \Bigg] ($C$=8) &
\Bigg[ \parbox[]{1.4cm}{256, 1$\times$1, 32\\ 32,  3$\times$3, 32\\ 32,  1$\times$1, 256} \Bigg] ($C$=8), $\times$2    

\\
\\

\midrule
 {Average Pool} & {Average Pool} \\
{100 fc, softmax} & {100 fc, softmax}\\

\bottomrule

\end{tabular}
}
\end{table}

%
%
%
%
%

\end{document}